\documentclass{article}
\usepackage{spconf,amsmath,graphicx}

\title{IMPROVING SPEECH RECOGNITION ERROR PREDICTION \\  FOR MODERN AND OFF-THE-SHELF SPEECH RECOGNIZERS}
\name{Prashant Serai \qquad Peidong Wang \qquad Eric Fosler-Lussier }
\address{The Ohio State University \\ \textit{Department of Computer Science \& Engineering}}
\begin{document}
%
\maketitle
\begin{abstract}
Modeling the errors of a speech recognizer can help simulate errorful recognized speech data from plain text, which has proven useful for tasks like discriminative language modeling, improving robustness of NLP systems, where limited or even no audio data is available at train time. Previous work typically considered replicating behavior of GMM-HMM based systems, but the behavior of more modern posterior-based neural network acoustic models is not the same and requires adjustments to the error prediction model.  In this work, we extend a prior phonetic confusion based model for predicting speech recognition errors in two ways: first, we introduce a sampling-based paradigm that better simulates the behavior of a posterior-based acoustic model.  Second, we investigate replacing the confusion matrix with a sequence-to-sequence model in order to introduce context dependency into the prediction.  We evaluate the error predictors in two ways: first by predicting the errors made by a Switchboard ASR system on unseen data (Fisher), and then using that same predictor to estimate the behavior of an unrelated cloud-based ASR system on a novel task. Sampling greatly improves predictive accuracy within a 100-guess paradigm, while the sequence model performs similarly to the confusion matrix.

\end{abstract}
\begin{keywords}
Speech Recognition, Error Prediction, Low Resource, Sequence to Sequence Neural Networks, Simulated ASR Errors
\end{keywords}
\section{Introduction}
\label{sec:intro}

Automatic Speech Recognition (ASR) is proliferating quickly, with a variety of applications having speech as an input modality. Yet, application specific audio data is often a scarce resource, and models trained on text data are widely being paired with cloud based speech recognition services. The nature of speech recognized text can be different from typed text data, notably in the nature of errors i.e. typos vs. speech recognition errors. Prior work has shown that given text data it is possible to simulate the recognition errors that might occur if the text were to be spoken, and the benefit of such simulated data especially in the absence of in application specific ASR data.

Fosler-Lussier et al. described a Weighted Finite State Transducer (WFST) framework that models word errors in speech recognition by measuring kinds of phonetic errors to build a phonetic confusion matrix \cite{fosler2005framework}. Anguita et al. looked inside the HMM-GMM acoustic model of the speech recognizer to directly determine phone distances and model errors \cite{anguita2005detection}. Jyothi and Fosler-Lussier combined the two aforementioned ideas and extended it to predict complete utterances of speech recognized text \cite{jyothi2009comparison}. Several works have used simulated ASR error data to do discriminative training and improve ASR performance \cite{sagae2012hallucinated,jyothi2010discriminative,kurata2011training}. Simulating ASR errors can also help with downstream tasks: Tsvetkov et al. incorporated knowledge of simulated ASR errors at train time, to improve the performance of a Machine Translation system in the face of real speech recognized data at test time \cite{tsvetkov2014augmenting}. 

There is not prior published work, to the best of our knowledge, that predicts errors for an ASR system using a neural network acoustic model, or for commercial off-the-shelf recognizers. With modern speech recognizers we can no longer determine phone distances by peeking into the acoustic model, so some of the enhanced confusion transducer based models \cite{anguita2005detection,jyothi2009comparison} are no longer applicable. Also, off-the-shelf speech recognizers do not afford much access to internals or information about them. 

However, there is a key flaw in the Fosler-Lussier et al. WFST model that implies a significant mismatch to posterior based models: the confusion matrix is used {\em directly} for prediction, and does not exhibit the peaky behavior of frame-level (or CTC-level) posterior estimates.  This mismatch creates a poor model when generating errors for neural net based systems.  The key insight in this paper is that we use the confusion matrix over all examples to sample {\em which} phones should appear in the output distribution, but provide a distribution that is much more peaky and ignores the smoother tails of the confusion matrix.  When sampling from the confusion matrix instead of composing with its FST form, we find that it better models the stochasticity of errors and confidence of neural network acoustic models. 

The standard confusion matrix in the WFST framework is context-independent, which does not reflect well the context-dependence of errors (for example, a canonical vowel next to /r/ will more likely be misidentified as an r-colored vowel than other vowels).  However, context-dependent confusion networks require substantial amounts of data to train and will be relatively sparse.  In order to model context dependence, we trained a neural Sequence to Sequence model, predicting errorful phone sequences from canonical sequences.  We can then either directly use the predictions, or decode using the WFST framework above.


In the next section, we describe the two models used in this experiment.  Section~\ref{sec:expsetup} details the experimental setup, followed in Section~\ref{sec:results} by evaluation both on a same-setting task (train on Switchboard, test on Fisher with the same recognizer) and a cross-setting task (train on Switchboard, test on a dialogue system using a different recognizer).

\begin{figure}[tb]

\begin{minipage}[b]{1.0\linewidth}
  \centering
  \centerline{\includegraphics[width=8.5cm]{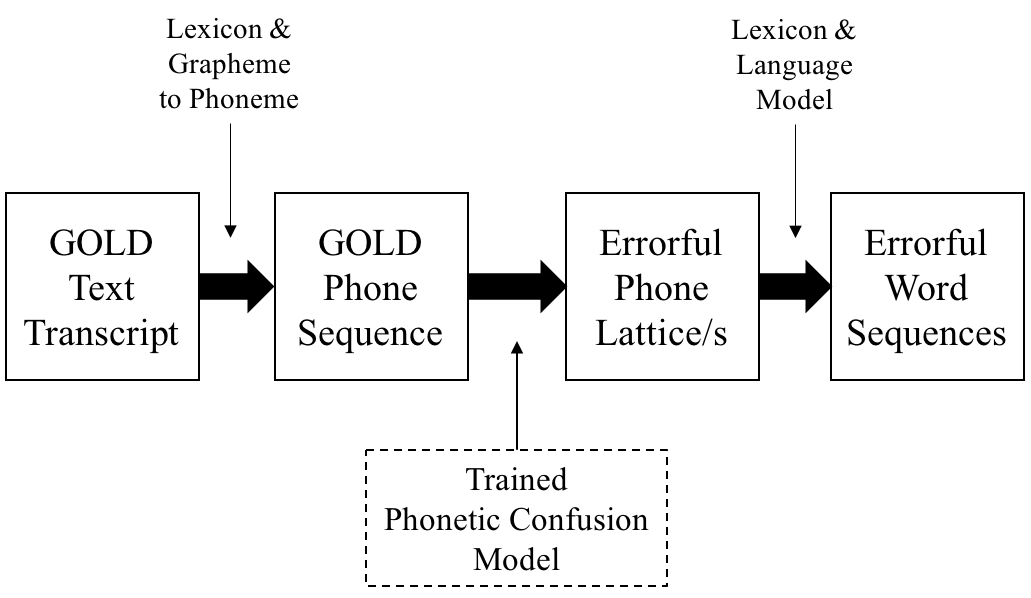}}
\end{minipage}
\caption{General pipeline for generating simulated errorful sequences: text transcripts are converted to phone sequences, which are converted to confusable phone lattices through the error prediction model.  These are then decoded by a WSFT model incorporating a lexicon and language model.}
\label{fig:pipeline}
\end{figure}

\section{Model Descriptions}
\label{sec:models}
Our general pipeline for going from regular text to text with simulated ASR errors is shown in Figure\ref{fig:pipeline}.
We convert the input word sequence to a phone sequence, simulate phonetic errors, and then decode back to a word sequence. We focus our attention on simulating the errors at the phonetic level.
\subsection{Confusion Matrix Based Models}
The WFST error prediction approach \cite{fosler2005framework,jyothi2009comparison} estimates an N-best list of word confusions $W_{\it conf}$ from an input word sequence $W$ through the following equation:
\[W_{\it conf} = W\:o\:P^{-1}\:o\:C\:o\:P\:o\:L\]
The words in the original text sequence are converted to phones using pronunciations from an inverted lexicon \(P^{-1}\), composed with a confusion matrix WFST \(C\), and is then decoded back into a simulated ASR transcript i.e. word sequence by composing a lexicon \(P\) and language model \(L\), which are also FSTs. The confusion matrix (ConfMat) transduces every phone at the input to a sequence of phones from length 0 (deletion) to 1 (no error/mutation) or more than 1 (insertion), and can be pre composed with the lexicon and language model for efficiency. For each original text sequence, once we have the final composed \(W_{\it conf}\) graph we calculate the N-best unique strings \cite{mohri2002efficient} to obtain N alternative word sequences to the input.  We vary the lexicon and grammar in the WFST predictor for each dataset.


Instead of directly using the confusion matrix for prediction, we can also sample the matrix to determine output sequence.  We convert the words in the original text sequence to phones, but then instead of composing with the confusion matrix WFST, for each phone in the input we sample without replacement, two options or alternatives for it from the ConfMat based on the probabilities with which they were observed to be confusible with the input phone. The most likely option for each input phone is typically the same phone itself, but the hope here is that over several iterations of sampling, the simulated recognizer will pick errors of various kinds over the input phone. We construct an FST by chaining these sampled alternatives, and weight the first sampled option with the weight of the most likely option, and weight the second sampled option with the weight of the second most likely option in the ConfMat, and then normalize. Finally, we compose with a decoding graph composed from a lexicon and language model as above and calculate the 1-best string for each sampled WFST per input sequence, and combine the various output word sequences obtained, ranking them by frequency of occurrence.


\subsection{Neural Sequence to Sequence Based Models}
We also experimented with context-dependent prediction of the errors using a sequence-to-sequence model.
To model phonetic confusions with information about the context, we use a 2-layer 128-unit recurrent neural Sequence to Sequence Model (Seq2Seq) \cite{sutskever2014sequence} with attention \cite{bahdanau2014neural}. We feed the model with the phonetic transcript of the true word sequence at the input, and at each time step of the input we provide an additional one hot vector containing one of five cues representing different kinds of errors (no error, mutation, deletion, insertion of one additional phone, insertion of more than one additional phones). At train time, we align the input and output phone sequence using the same technique as the confusion matrix based system, to determine what kind of error is being made (or not made) for each time step of the input. At test time, we randomly sample to select one of the five aforementioned cues for each timestep of the input, from a collapsed version of the confusion matrix that holds information about the frequencies of these five kinds of errors for each input phone. At test time, the probability distributions at the output of the Seq2Seq model are then converted into WFSTs, by selecting three phones at each time step, and a softmax function with a temperature \(\tau=10\) is applied as shown below.
\[P_t(a) = \frac{\exp(q_t(a)/\tau)}{\sum_{i=1}^n\exp(q_t(i)/\tau)}\]

Once we have the WFSTs, we compose them with a decoding graph as in the case of the confusion matrix model, except the decoding graph is augmented to absorb (with a small cost) any number of ``EOS'' (End of Sequence) symbols at the end of phone sequences being translated into word sequences.  Finally, 5-best word sequences are calculated for each input sample, and combined to produce K alternatives per input text sequence.
%

Besides sampling for cues at the input of the Seq2Seq model, we also explore the effect of additional sampling from the output probability distribution produced by the Seq2Seq model. Similar to the sampling from the distributions of confusion matrix, instead of directly producing WFSTs from the output of the Neural network, we generate multiple sampled WFSTs. Owing to existence of the additional ``EOS'' symbol, we pick three rather than two alternatives at each time step, and then decode the WFSTs thus sampled with the same EOS augmented decoding graph as above.

\section{Experimental Setup}
\label{sec:expsetup}
\subsection{Data}
We use the Kaldi Switchboard recipe to train a Deep Neural Network acoustic model on the Switchboard corpus. A sMBR criterion is used during training and decoding proceeds with a trigram grammar \cite{vesely2013sequence}.
We use this recognizer to transcribe speech data from the Fisher corpus \cite{cieri2004fisher}, containing about 1.8 million utterances with a word error rate of roughly 30\%. We use the speech recognized text from Fisher paired with gold text as the training set for all our Models, holding out a validation set of 100 utterances for the tuning of hyperparameters.
We tested our models on two kinds of data. Firstly, we predicted errorful transcripts for held out data from the Fisher corpus, which was recognized by the aforementioned speech recognizer that we trained. This was a set of 500 utterances containing 504 error chunks across all recognized speech.  The WFST prediction module uses the standard Switchboard lexicon and grammar used in the recognizer.

Secondly, we predicted errorful transcripts for data from the Virtual Patient project \cite{jin2017}, where volunteers read out text data from doctor trainees querying a patient avatar.  The recorded speech was recognized using a cloud based ASR service treated as a black box \cite{stiff2018phonecnn}. This was a set of 756 utterances containing 258 error chunks across all recognized speech, and the word error rate was slightly over 10\%.  As there is a vocabulary mismatch between Switchboard and the Virtual Patient, but we did not want to inform the error prediction system of the Virtual Patient vocabulary, we extended the WFST lexicon and language model by leveraging models from the EESEN Offline Transcriber \cite{miao2015eesen}, which uses a pruned 3-gram language model provided by Cantab Research \cite{cantab}  trained on TED-LIUM data \cite{rousseau2014enhancing}.

\subsection{Training Details}

To train the confusion matrix models, we first convert the gold and speech recognized transcripts to phone sequences, and align them using a phonetic distance based dynamic programming algorithm \cite{fosler2005framework}. The alignment is done in such a way that each input phone (e.g. /s/) is paired with a sequence of phones of length 0 (deletion, /s/:$\epsilon$), 1 (no error or mutation, /s/:/s/ or /s/:/z/) or more (insertion, /s/:/st/).
For each possible input phone, we count frequencies of various such ``alternative'' phone sequences, and normalize them into probabilities. This gives us our confusion matrix for composing or sampling.

For the training the Sequence to Sequence (Seq2Seq) based phonetic confusion model, we start with unaligned pairs of gold and errorful phone sequences, and train it to minimize the cross entropy between the model predictions and the ground truth (i.e., the errorful phone sequence). In our experiments, we found that when we directly used the ground truth sequence, which is a one hot distribution at each time step, the model predictions would be very peaky and not provide much diversity for decoding. To allow the model to learn to produce more diversity at the output, we smoothed the ground truth sequence with alternatives from the confusion matrix at train time:
\[Y_{\it smooth} = \beta * Y_{\it original} + (1-\beta) * C_{11}[y]\]
\noindent where \(y\) is the original phone label and \(Y_{\it original}\) is the one hot probability distribution corresponding to it, and \(Y_{\it smooth}\) is the smoothed probability distribution. \(C_{11}\) is a reduced version of the confusion matrix that only has the alternatives of length 1 (mutation or no error), and \(\beta\) is the smoothing factor (we use value 0.8). Although \(C_{11}\) only captures mutation errors, we observed that on smoothing, the neural network was automatically learning to give meaningful weight to output phones at their adjacent timesteps as well. Finally, we use an Adam optimizer to minimize the cross entropy of the model predictions with the smoothed ground truth across minibatches of 64 examples at every training step.

\section{Evaluation and Results}
\label{sec:results}
Following prior work, we use two metrics to evaluate the effectiveness of our models in simulating ASR errors. The first metric measures the percentage of real test set Error Chunks recalled in a set of ``K best'' simulated speech recognized utterances for each gold word sequence. The error chunks are again determined by aligning the gold word sequence with the errorful word sequence and removing the longest common subsequence. For example, if the gold sequence is ``do you take any other medications except for the tylenol for pain'' and the errorful sequence is ``you take any other medicine cations except for the tylenol for pain,'' the error chunks would be the pairs \(\{medications:medicine\:cations\}\) and \(\{do:\:\:\}\). Our detection of error chunks is strict --- for an error chunk to qualify as predicted, the words adjacent to the predicted error chunk should be error-free.

The second metric measures the percentage of times the complete test set utterance is recalled in a set of ``K best'' simulated utterances for each gold text sequence (including error-free test sequences). We aimed to produce 100 unique simulated speech recognized utterances for each gold word sequence, so for both of these metrics, we evaluate the performance at K=100.

Note that these are both hard metrics since the possibilities of various kinds of errors is quite endless, and no matter the plausibility of the output, the metrics only give credit when the utterance/error chunk exactly matches what was produced.

\begin{table}[tb]
\begin{tabular}{|p{0.48\linewidth}|p{0.16\linewidth}|p{0.18\linewidth}|}
	\hline
	Model & Error Chunks Predicted & Complete Utterances Predicted\\
	\hline
    ConfMat Direct decoding & 14.9\% & 39.2\%\\
	\hline
    ConfMat Sampled decoding & \textbf{25.6}\% & 38.8\%\\
    \hline
    Seq2Seq Direct decoding & 23.8\% & 38.2\%\\
    \hline
    Seq2Seq Sampled decoding & 23.0\% & 37.8\%\\
    \hline
    Seq2Seq Direct (K=50) + ConfMat Sampled (K=50) & 23.8\% & \textbf{43.4}\%\\
	\hline
\end{tabular}
\caption{Evaluation on unseen Fisher corpus recognition data from the same recognizer}
\label{table-fisher}
\end{table}

Table \ref{table-fisher} shows the results on the held out set from the Fisher corpus for all of the models tried. The sampled decoding on the confusion matrix greatly improves over the direct baseline model from \cite{fosler2005framework} in terms of real error chunks predicted, maintaining comparable performance on the complete utterance prediction metric. The neural Seq2Seq models also provide a significant improvement over the baseline in terms of error chunks predicted, although the sampling on the confusion matrix still performed better on that metric. Since the neural Seq2Seq model has the capacity to model errors in a context dependent manner, we wanted to see if it was learning something different from the confusion matrix models. We combined half the number of utterances from the best ConfMat approach and the best Seq2Seq approach each, and looked at the metrics for that.  The number of complete utterances recalled was significantly higher than either of the approaches being combined, suggesting that the Seq2Seq model might be learning things that the ConfMat model is not able to capture.

\begin{table}[tb]
\begin{tabular}{|p{0.48\linewidth}|p{0.16\linewidth}|p{0.18\linewidth}|}
	\hline
	Model & Error Chunks Predicted & Complete Utterances Predicted\\
	\hline
    ConfMat Direct Decoding & 8.5\% & 66.9\%\\
	\hline
    ConfMat Sampled Decoding & 36.4\% & 72.4\%\\
    \hline
    Seq2Seq Direct Decoding & 20.2\% & 68.3\%\\
    \hline
    Seq2Seq Sampled Decoding & 16.7\% & 67.7\%\\
    \hline
    Seq2Seq Direct (K=50) + ConfMat Sampled (K=50) & 34.9\% & 71.8\%\\
	\hline
\end{tabular}
\caption{Evaluation on Virtual Patient data from a cloud-based ASR service}
\label{table-vp}
\end{table}

Table \ref{table-vp} shows the results on predicting recognition errors made by the cloud based ASR service. We see that the baseline ConfMat with direct decoding fares even more poorly here. Our proposed sampling approach yields major gains in terms of both metrics, notably recalling almost four times the number of error chunks as the the baseline. The Seq2Seq model does better than the baseline, but does not do as well as the confusion matrix based system with sampled decoding on either metric.

The confusion matrix based system with sampled decoding has also proven to help with a downstream task:  Stiff et al. \cite{stiff2018phonecnn} adapted a chatbot answer prediction system to work better with speech input.  The 
error predictor was used to simulate speech errors which were sampled by the classification system during training.  The addition of the sampled errorful data to the training regime improved the spoken interpretation accuracy modestly but consistently across several training conditions (e.g. from 67.6\%  to 68.3\% accuracy in the best performing system). 

\section{Conclusions and Future Work}
\label{sec:conc}
We show that the sampling based paradigm greatly improves the error prediction performance of the confusion matrix system. We observe that while the Seq2Seq confusion model might be learning to predict errors in a context dependent manner, the method does not generalize well across corpora, and needs further work. We think the Seq2Seq model may benefit from enhancements such as a more robust generator for the error types, multi-head attention, scheduled sampling, and beam search decoding.

\section{Acknowledgements}
\label{sec:ack}
This material is based upon work supported by the National Science Foundation under Grant No. 1618336.  We thank Adam Stiff for sharing the paired text and speech recognized data from the Virtual Patient project for our experiments.



\bibliographystyle{IEEEbib}
\bibliography{strings,refs}

\end{document}